\documentclass{article}
\usepackage{spconf,amsmath,graphicx}
\usepackage{cite}
\usepackage{color}

\graphicspath{{./figs/}}

\title{Hybrid deep neural networks for all-cause Mortality Prediction from LDCT Images}
\twoauthors
  {Pingkun Yan, Hengtao Guo, Ge Wang}
	{Department of Biomedical Engineering\\
	Rensselaer Polytechnic Institute\\
	Troy, NY 12180, USA}
  {Ruben De Man, Mannudeep K. Kalra}
	{Department of Radiology\\ 
	Massachusetts General Hospital\\
	Boston, MA 02114, USA}
\begin{document}
%
\maketitle
\begin{abstract}
Known for its high morbidity and mortality rates, lung cancer poses a significant threat to human health and well-being. However, the same population is also at high risk for other deadly diseases, such as cardiovascular disease. Since Low-Dose CT (LDCT) has been shown to significantly improve the lung cancer diagnosis accuracy, it will be very useful for clinical practice to predict the all-cause mortality for lung cancer patients to take corresponding actions. In this paper, we propose a deep learning based method, which takes both chest LDCT image patches and coronary artery calcification risk scores as input, for direct prediction of mortality risk of lung cancer subjects. The proposed method is called Hybrid Risk Network (HyRiskNet) for mortality risk prediction, which is an end-to-end framework utilizing hybrid imaging features, instead of completely relying on automatic feature extraction. Our work demonstrates the feasibility of using deep learning techniques for all-cause lung cancer mortality prediction from chest LDCT images. The experimental results show that the proposed HyRiskNet can achieve superior performance compared with the neural networks with only image input and with other traditional semi-automatic scoring methods. The study also indicates that radiologist defined features can well complement convolutional neural networks for more comprehensive feature extraction.
\end{abstract}

\section{Introduction}
\label{sec:intro}

Low-dose CT (LDCT) has been proven effective in lung cancer screening, where the National lung Screening Trial (NLST) observed 20\% decrease in lung cancer related mortality in at-risk subjects (55 to 74 years, 30 pack-year cigarette-smoking history) \cite{NLST_2011}. Chiles et~al. \cite{chiles_association_2015} show that coronary artery calcification (CAC) scored using three different methods is strongly associated with mortality in NLST. In a different study – the Dutch-Belgian Randomized Lung Cancer Screening Trial (NELSON), it was found that CAC can predict all-cause mortality and cardiovascular events on lung cancer screening LDCT \cite{jacobs_coronary_2012}.
Our previous work \cite{digumarthy_multifactorial_2018} also demonstrates significant difference in CAC scores between the survivor and non-survivor groups, which indicates that CAC can be used for predicting mortality risk of lung cancer patients. However, the existing risk quantification methods rely on directly counting the number of pixels that are above a certain threshold \cite{agatston_quantification_1990,callister_coronary_1998}.
Although the calcium size carries valuable information, it can also miss some key mortality indicators as presented in \cite{digumarthy_multifactorial_2018}.

In the past several years, machine learning – especially its subdomain of deep learning – has seen a series of breakthroughs \cite{krizhevsky_imagenet_2012}. With deep learning being a paradigm shifter, a large number of innovations have also been made in medicine, ranging from medical image processing to health record analysis. 
%
%
Deep learning has also been applied for automatic calcium scoring in chest LDCT \cite{lessmann_automatic_2018}, which reports that deep neural networks can be used for measuring the size of CAC from LDCT and different filters used during reconstruction may influence the quantification results. However, to train such a network, it requires manually labeling the area of calcification from images. Due to the significant efforts involved in labeling, only a small number of images can be marked and then used for algorithm training, which may well limit the performance. More importantly, those methods cannot extract other imaging markers for predicting mortality risk. Very recently, van~Velzen et al. \cite{van_velzen_direct_2018} proposed to use deep learning for cardiovascular mortality from LDCT images.

In this paper, we propose a method using convolutional neural networks (CNNs) with hybrid input for directly predicting all-cause mortality risk of lung cancer patients from their LDCT images. The seminal architecture of deep residual neural network (ResNet) \cite{he_deep_2016} is adopted for extracting image features, due to its effectiveness in training very deep networks for extracting high-level representative imaging markers. 
With that, we aim to extract not only calcification information but also other auxiliary features for better prediction. In the same time, since various pre-defined imaging markers have been well recognized as indication of mortality risk, it will be a significant waste if we just throw those features away. Therefore, the proposed HyRiskNet also takes CAC scores as an additional input to complement the ResNet for more accurate prediction.
Our experimental results show that the proposed method can obtain much more accurate results on mortality risk quantification than the traditional methods.

The presented work has three main contributions. First, instead of simply measuring the extent of CAC as a surrogate index, deep neural networks are used for direct lung cancer patient all-cause mortality risk prediction.
Second, various deep CNNs are compared and validated for the challenging task to understand their performance.
Last, our results demonstrate that the deep learning methods, when complemented with pre-defined imaging marker, can achieve even better results than being used alone.

The rest of this paper is organized as follows. Section~\ref{sec:materials} provides the details of the used materials in our work. The proposed methods are presented in Section~\ref{sec:methods}. The experimental results and discussions are given in Section~\ref{sec:results}. Finally, Section~\ref{sec:conclusions} draws conclusions about the presented work.

\begin{figure*}[tbp]
	\label{fig:risknet}
	\centering
	\includegraphics[width=.75\textwidth]{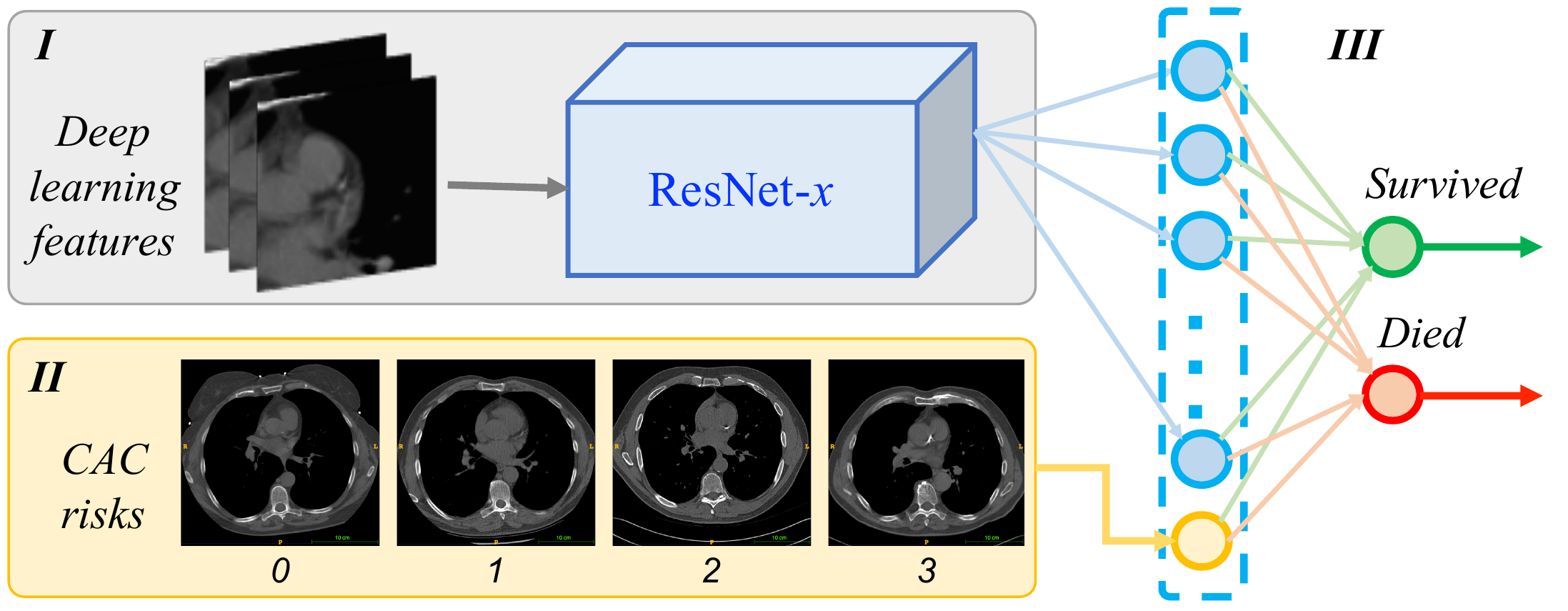}
	\caption{Overview of the proposed HyRiskNet. RiskNet uses only features in blue nodes for risk prediction, while HyRiskNet employs all feature nodes.}
\end{figure*}

\section{Materials}
\label{sec:materials}

In this paper, all the study data were acquired by the National Lung Screening Trial (NLST) \cite{chin_screening_2015}, which are managed by the National Cancer Institute Cancer Data Access System. In this large scale clinical trial, NLST compared LDCT with the chest radiography for lung cancer screening in more than 50,000 current or former smokers who met the various inclusion criteria. In our work, following the same protocol used in \cite{digumarthy_multifactorial_2018}, 180 subjects (90 survivors, 90 non-survivors) were selected for the study, each group consisting of 49 subjects with stage I, 19 subjects with stage II, and 22 subjects with stage III lung cancers. Each patient went through three LDCT exams, of which the first LDCT scan of each patient is used in this study.

CAC qualification is considered as a baseline benchmark for performance comparison. To achieve that, two radiologists graded CAC in all 180 CT examinations in a blinded and randomized manner. CAC was graded on a 4-point scale (0 = none, 1 = minimal, 2 = moderate, and 3 = severe) based on a prior publication on subjective evaluation of CAC on LDCT for lung cancer screening \cite{chiles_association_2015}. The radiologists were provided with examples of each CAC grade to use as a reference.

The Agatston score and risk category \cite{agatston_quantification_1990}, the original volume score \cite{callister_coronary_1998}, and the square root of volume score in order to decrease the variation \cite{hokanson_evaluating_2004} were measured in images and reported. The image patches containing the heart region defined by a center location with size of 65$\times$65 pixels from 5 slices were analyzed by a Python program. Since the NLST scans were performed at 120-140 kVp, a threshold pixel value (130 HU) was applied to detect pixels above the threshold being considered as calcification.

\section{Methods}
\label{sec:methods}

In this section, we present our proposed method for mortality risk prediction using LDCT images. Instead of having only image input, we propose a hybrid neural network to also take CAC score as another input to better use radiologist knowledge for improved risk prediction. The proposed deep learning method is coined as Hybrid-Risk-Net and the overall architecture of the proposed network is shown in Fig.~\ref{fig:risknet}. Details of the proposed method are provided as follows.

\subsection{Network Design}

As it can be seen in Fig.~\ref{fig:risknet}, the proposed HyRiskNet consists of three major segments (I, II and III).
The first segment is to compute the deep learning features from input LDCT cardiac image patches. The deep residual network (ResNet) proposed by He et~al.~\cite{he_deep_2016}, which is one of the top performing deep CNNs in various computer vision tasks, has been adopted as the backbone of our method for this difficult task. By using only the convolutional layers of ResNet, image features can be extracted by ResNet-$x$, where $x$ denotes the depth of the network. 
At the end of the convolutional layers, 512 features are extracted by ResNet-18 and 34, and 2048 features are ResNet-50, 101 and 152.
The features can be used for mortality risk prediction through a fully connected layer as shown in Fig.~\ref{fig:risknet}, which is referred as RiskNet-$x$ herein.
The second part uses CAC scores as input to take advantage of features defined using medical domain knowledge. The hybrid features are then fused together by concatenation and fed to a fully connected layer for predicting if a subject will survive or die at the end of the trial. The entire network with all three components together is called HyRiskNet-$x$.

\subsection{Implementation Details}

The proposed HyRiskNet was implemented using the open-source PyTorch library \cite{pytorch} in Python programming language.
To efficiently use the well defined ResNet-$x$ networks, 3-channel images are constructed by stacking 3 neighboring slices together. Training of the network is completed in two stages. In the first stage, three different training strategies are used for training RiskNet-$x$, which uses only LDCT image patches. Strategy one is to directly use the ResNet-$x$ pre-trained on ImageNet data for feature extraction. Strategy two is to fine-tune the pre-trained ResNet-$x$ on our data using Adam optimizer with learning rate of 0.00001. The last strategy is to train the RiskNet-$x$ from scratch using Adam optimizer with the initial learning rate of 0.0001, which decays by 0.9 after every 5 epochs. Once the RiskNet-$x$ is trained, in the second stage, either CAC scores or Agatston scores are added in for training the full HyRiskNet-$x$, and both are normalized to the scale between -1 and 1.


Data augmentation operations of random cropping and scaling are used for training the networks. The image patches in the size 161$\times$161 pixels are cropped from LDCT images centered at the same coordinates clicked by radiologists for other CAC scoring methods. The input patches are randomly cropped with scaling ratio between 0.6 and 0.8 and resized to 224$\times$224 pixels for network input. All the image patches are then normalized by subtracting the mean intensities and being divided by the standard deviation as 3-channel images.

\section{Results and Discussions}
\label{sec:results}

In this section, we present our experimental results and the corresponding discussions.

\subsection{Performance Evaluation}

A ten-fold cross validation scheme was applied over our dataset for evaluating the performance of the proposed method. Since it essentially aims to perform binary classification of subjects into either ``survivor'' or ``nonsurvivor'', Receiver Operating Characteristic (ROC) curves are drawn to demonstrate the performance. Area Under the Curve (AUC) scores are used to compare the performance of different methods.

We first evaluated different ResNet training strategies for the targeted task, including directly using features from the convolutional components of ResNet pre-trained on ImageNet, fine-tuning pre-trained ResNet on our data, and training the entire network from scratch. Our experiments show that the network trained with random initialization works the best. It is an indication that the significant difference between natural images and the LDCT images may make the pre-trained network not very useful for our task.

\begin{table}[htbp]
\centering
\caption{Mean AUC and standard deviation (Std. Dev.) of RiskNet and HyRiskNet in different depths.}
\label{tab:RiskNets}
 \begin{tabular}{ l | c }
 	\hline
 	Network		& AUC Mean $\pm$ Std. Dev. \\ \hline
 	RiskNet-18	& 0.67 $\pm$ 0.09	\\
 	\textbf{RiskNet-34}	& \textbf{0.70 $\pm$ 0.08}	\\
 	RiskNet-50	& 0.68 $\pm$ 0.09	\\
 	RiskNet-101	& 0.66 $\pm$ 0.10	\\
 	RiskNet-152	& 0.68 $\pm$ 0.07	\\ \hline
 	HyRiskNet-18	& 0.69 $\pm$ 0.13	\\
 	\textbf{HyRiskNet-34}	& \textbf{0.75 $\pm$ 0.14}	\\
 	HyRiskNet-50	& 0.70 $\pm$ 0.10	\\
 	HyRiskNet-101	& 0.67 $\pm$ 0.11	\\
 	HyRiskNet-152	& 0.72 $\pm$ 0.09	\\ \hline
 \end{tabular}
\end{table}

Table~\ref{tab:RiskNets} shows the performance of the designed networks at different depth, where we include both RiskNet and HyRiskNet for comparison. As it can be seen, HyRiskNet works generally better than RiskNet with the same depth, which suggests that the added CAC scores actually help to better predict the mortality risk. We also compared the effects of network depth. As shown in Table~\ref{tab:RiskNets}, for both RiskNet and HyRiskNet, the architecture with 34 layers provides the best performance. This may be because networks with 18 layers may not be powerful enough to extract the representative features from LDCT images. On the other hand, networks with 50 or more layers have large number of parameters, which may not get fully trained given our relatively small dataset. It interesting to see that both RiskNet-152 and HyRiskNet-152 achieve the second best results in their own groups, which is worth of more investigation in our future work with larger datasets.

\subsection{Effects of Risk Scores}

\begin{figure}[tb]
	\centering
	\includegraphics[width=\columnwidth]{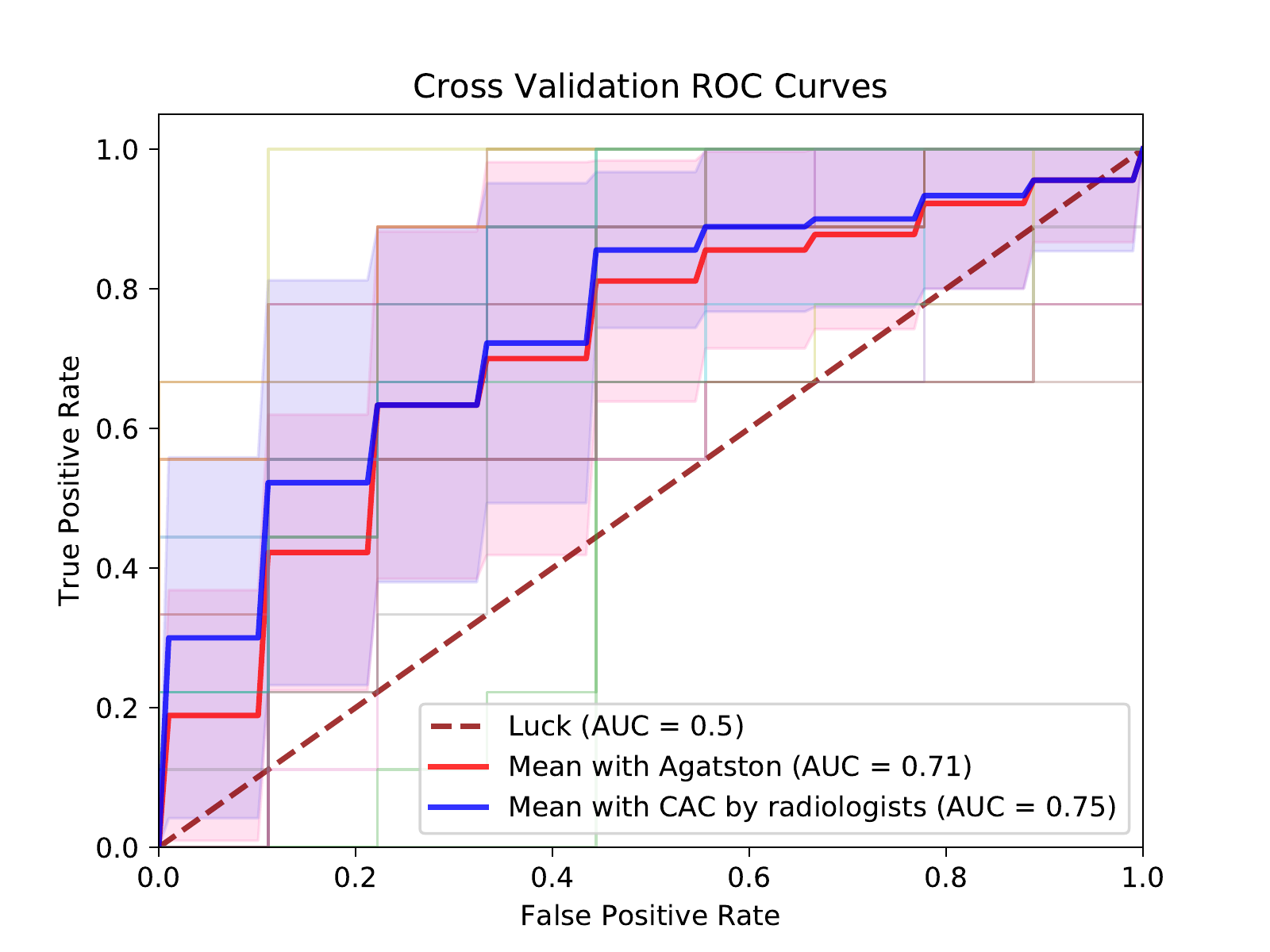}
	\caption{Ten-fold cross validation ROC curves of HyRiskNet-34, with CAC risk scores from radiologist by subjective evaluation and Agatston evaluation, respectively.}
\label{fig:cross_validation}
\end{figure}

We further evaluated the effects of risk scores by comparing the performance of HyRiskNet using CAC risk scores obtained either subjectively by the radiologists as described in Section~\ref{sec:materials} or objectively through Agatston method. The ten-fold cross validation results are shown in Fig.~\ref{fig:cross_validation} using ROC curves, which indicates that on average HyRiskNet-34 always works better when supplied with CAC risk scores obtained by radiologists with mean AUC=0.75. In contrast, when being provided with Agatston scores, the mean AUC value of HyRiskNet-34 dropped to 0.71.

\subsection{Comparison against other scoring methods}

\begin{figure}[tb]
	\centering
	\includegraphics[width=\columnwidth]{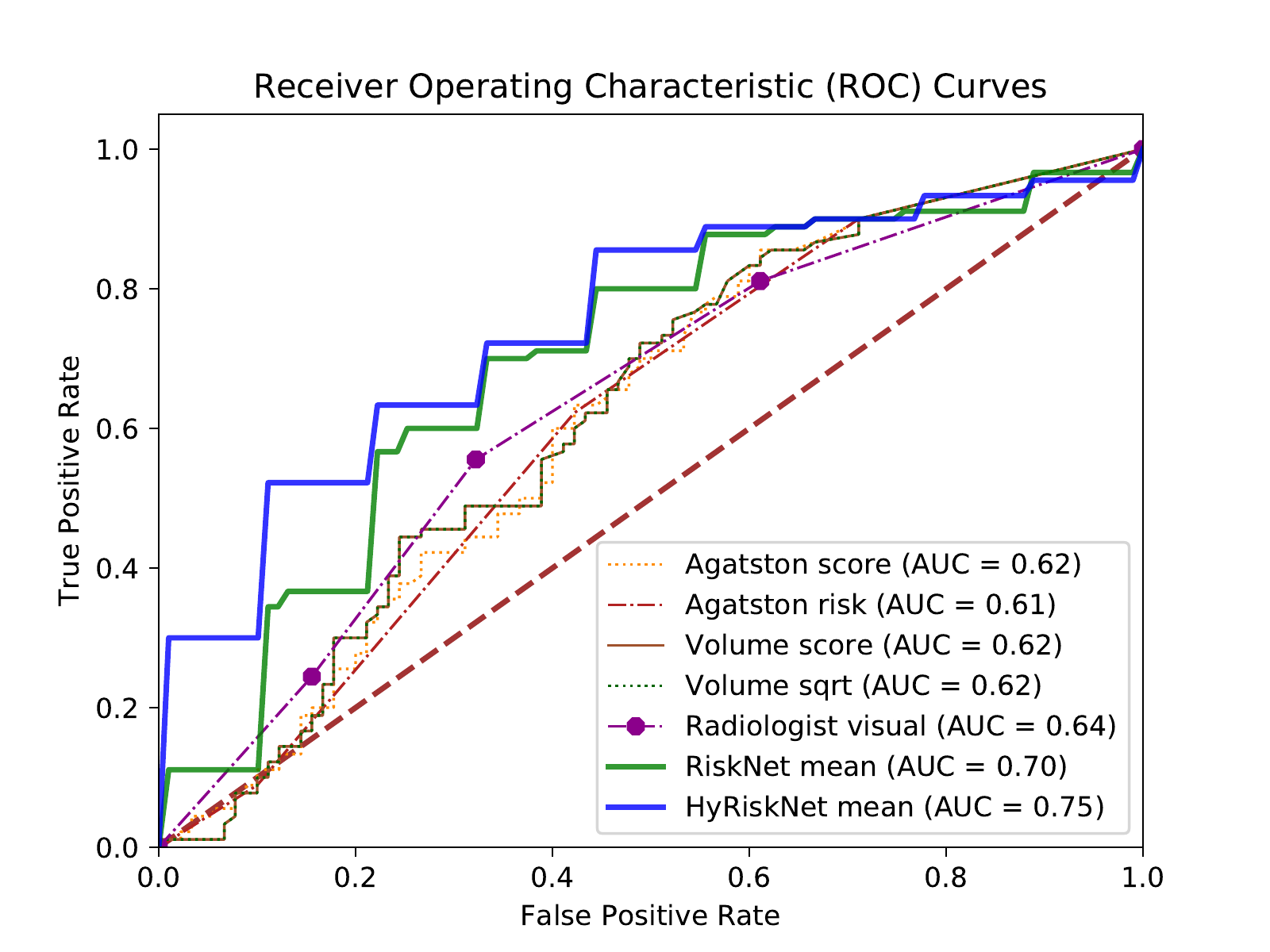}
	\caption{Performance comparison of various methods on all-cause mortality risk prediction.}
\label{fig:compare}
\end{figure}

The developed HyRiskNet is then compared against other scoring methods for further validation. The results are shown in Fig.~\ref{fig:compare}. It can be seen that the traditional semi-automatic methods, such as Agatston score \cite{agatston_quantification_1990}, Agatston risk, volume score \cite{callister_coronary_1998}, and square root of volume score \cite{hokanson_evaluating_2004}, perform similarly and the mean AUC values are all at 0.61 or 0.62, slightly better than random guess. It is very interesting to see that the visual inspection of CAC by radiologists outperforms the semi-automatic CAC scoring methods with AUC=0.64. This suggests that some information about the condition of cardiovascular vessels is not captured by those scoring methods, but has been taken into account by the radiologists.

The significant performance improvement comes from the proposed CNN based methods. Fig.~\ref{fig:compare} indicates that CNN can be used for extracting and quantifying features in cardiac patches from chest LDCT images for all-cause mortality prediction. RiskNet-34, which uses image features alone, outperforms all the CAC based scoring methods including subjective scoring by the radiologists with mean AUC of 0.70. When combining deep learning extracted features and radiologist risk scores together, the proposed HyRiskNet-34 achieves the best performance with AUC=0.75, which improves the prediction performance by 17.2\% and 7.1\%  over radiologist inspection and RiskNet, respectively.

\subsection{Discussions}

The results from this study are consistent with our previous work in \cite{digumarthy_multifactorial_2018}, where imaging markers such as muscle mass and fat attenuation were found to be associated with the chance of survival. Our work demonstrates that more imaging information can be extracted by using deep learning for mortality risk prediction. It remains to be analyzed what are those features extracted by deep learning and the importance of their role.

\section{Conclusion}
\label{sec:conclusions}

In this paper, we showed that CNN based feature extraction method can be used together with radiologist defined image feature for all-cause mortality from low-dose chest CT images. Superior performance has been achieved by the use of deep learning, compared to other traditional manual or semi-automatic approaches.
In our future work, we will investigate the features extracted by deep learning for better understanding of the approach. That may also help us to design better neural networks to further improve the prediction performance.

\section{Acknowledgments}

The authors thank the National Cancer Institute for access to NCI's data collected by the National Lung Screening Trial. The statements contained herein are solely those of the authors and do not represent or imply concurrence or endorsement by NCI.
The authors would also like to thank NVIDIA Corporation for the donation of the Titan Xp GPU used for this research.


\bibliographystyle{IEEEbib}
\bibliography{refs}

\end{document}